\documentclass[conference]{IEEEtran}
\usepackage{blindtext, graphicx}
\usepackage{subcaption}
\usepackage{amsmath}
\usepackage{color}
\usepackage{colortbl}
\usepackage{graphicx}
\usepackage{acronym}
\usepackage[T1]{fontenc}
\usepackage{etoolbox}
\ifCLASSINFOpdf
\else
\fi

\usepackage[noadjust]{cite}

\acrodef{lstm}		[\textsc{LSTM\xspace}]				{Long Short-Term Memory Neural Networks}
\acrodef{rnn}		[\textsc{RNN\xspace}]				{Recurrent Neural Networks}
\acrodef{dip}		[\textsc{DIP\xspace}]				{Document Image Processing Pipeline}
\acrodef{dipp}		[\textsc{DIPP\xspace}]				{Document Image Processing Pipeline}

\acrodef{cnn}		[\textsc{CNN\xspace}]				{Convolutional Neural Networks}

\acrodef{relu}		[\textsc{ReLUs\xspace}]				{Rectified Linear Units}

\IEEEoverridecommandlockouts

\acrodef{das}[\textsc{DAS}\xspace]{Document Analysis System}
\acrodef{ann}[\textsc{ANN}\xspace]{Artificial Neural Network}
\acrodef{cnn}[\textsc{CNN}\xspace]{Convolutional Neural Network}
\acrodef{fcn}[\textsc{FCN}\xspace]{Fully Connected Network}
\acrodef{gpu}[\textsc{GPU}\xspace]{Graphics Processing Unit}
\acrodef{svm}[\textsc{SVM\xspace}]{Support Vector Machine}
\acrodef{pca}[\textsc{PCA\xspace}]{Principal Component Analysis}
\acrodef{cc}[\textsc{CC\xspace}]{Connected Component}
\acrodef{bvlc}[\textsc{BVLC\xspace}]{Berkeley Vision and Learning Center}
\acrodef{sf}[\textit{smart FIX}]{\textit{smart \textbf{F}or \textbf{I}nformation e\textbf{X}traction}}
\acrodef{dfki}[\textsc{DFKI\xspace}]{German Research Center for Artificial Intelligence}
\acrodef{gt}[\textsc{GT}\xspace]{Ground Truth}
\acrodef{fp}[\textsc{$FP$}\xspace]{False Positives}
\acrodef{fn}[\textsc{$FN$\xspace}]{False Negatives}	
\acrodef{tp}[\textsc{$TP$\xspace}]{True Positives}
\acrodef{tn}[\textsc{$TN$\xspace}]{True Negatives}
\acrodef{ocr}[\textsc{OCR\xspace}]{Optical Character Recognition}
\acrodef{pkv}[\textsc{PKV\xspace}]{Private Health Insurance}

\newcommand*{\eg}		{e.g.\ }

\newcommand*{\cf}		{cf.\ }
\newcommand*{\sota}		{state-of-the-art\ }

\begin{document}


\title{Cutting the Error by Half: Investigation of Very Deep CNN and Advanced Training Strategies for Document Image Classification}

\makeatletter

\def\footnoterule{\relax%
  \kern-5pt
  \hbox to \columnwidth{\hfill\vrule width 0.5\columnwidth height 0.4pt\hfill}
  \kern4.6pt}

\makeatother



  

\author{
    \IEEEauthorblockN{ Muhammad Zeshan Afzal\IEEEauthorrefmark{1}\IEEEauthorrefmark{2}\IEEEauthorrefmark{4}\thanks{* These authors contributed equally to this work},
    Andreas K\"olsch\IEEEauthorrefmark{1}\IEEEauthorrefmark{2},
    Sheraz Ahmed\IEEEauthorrefmark{3},
    Marcus Liwicki\IEEEauthorrefmark{2}\IEEEauthorrefmark{5}}

    \IEEEauthorblockA{
        afzal@iupr.com,
        a\_koelsch12@cs.uni-kl.de,
        sheraz.ahmed@dfki.de,
        marcus.liwicki@unifr.ch
    }

    \IEEEauthorblockA{\IEEEauthorrefmark{2}MindGarage, University of Kaiserslautern, Germany}

    \IEEEauthorblockA{\IEEEauthorrefmark{3}DFKI, Kaiserslautern, Germany}

    \IEEEauthorblockA{\IEEEauthorrefmark{4}Insiders Technologies GmbH, Kaiserslautern, Germany}

    \IEEEauthorblockA{\IEEEauthorrefmark{5}University of Fribourg, Switzerland}

}


\maketitle

\IEEEpeerreviewmaketitle
\begin{abstract}

We present an exhaustive investigation of recent Deep Learning architectures, algorithms, and strategies for the task of document image classification to finally reduce the error by more than half.
Existing approaches, such as the DeepDocClassifier, apply standard Convolutional Network architectures with transfer learning from the object recognition domain.
The contribution of the paper is threefold:
First, it investigates recently introduced very deep neural network architectures (GoogLeNet, VGG, ResNet) using transfer learning (from real images). 
Second, it proposes transfer learning from a huge set of document images, i.e. $400,000$ documents. 
Third, it analyzes the impact of the amount of training data (document images) and other parameters to the classification abilities. 
We use two datasets, the Tobacco-3482 and the large-scale RVL-CDIP dataset.
We achieve an accuracy of $91.13\,\%$ for the Tobacco-3482 dataset while earlier approaches reach only $77.6\,\%$. Thus, a relative error reduction of more than $60\,\%$ is achieved. For the large dataset RVL-CDIP, an accuracy of $90.97\,\%$ is achieved, corresponding to a relative error reduction of $11.5\,\%$.



\end{abstract}



\section{Introduction}

An important step of the \ac{dipp} is the classification of the documents. 
An early classification of documents helps to process the subsequent processes in \ac{dipp} such as information extraction, text recognition etc~\cite{doclass_Dengel95}. Due to its fundamental importance, this area has been explored extensively. 
Earlier methods that have been dealing with document classification focused mainly on either exploiting the structural similarity constraints~\cite{doclass_Byun2000, doclass_shin} or extracting features from the documents that may be able to help for  document classification~\cite{doclass_Kumar12, doclass_Chen12, doclass_Kumar14}.
Some of the methods  combine both of the features~\cite{Collins-thompson02aclustering-based}.


\begin{figure}
\begin{subfigure}{\linewidth}
  \centering
    \includegraphics[width=47px, height=34px]{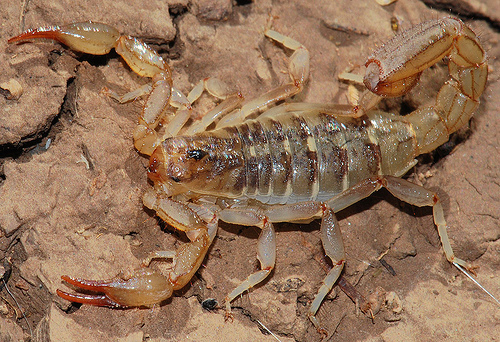}
    \includegraphics[width=47px, height=34px]{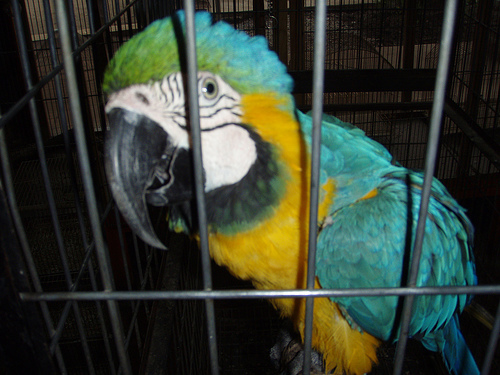}
    \includegraphics[width=47px, height=34px]{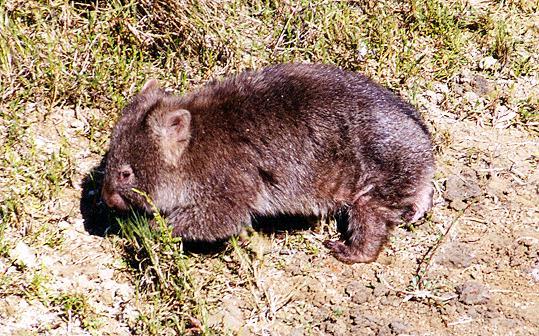}
    \includegraphics[width=47px, height=34px]{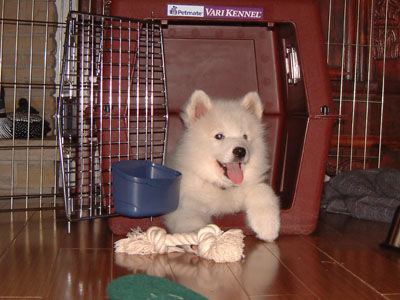}
    \includegraphics[width=47px, height=34px]{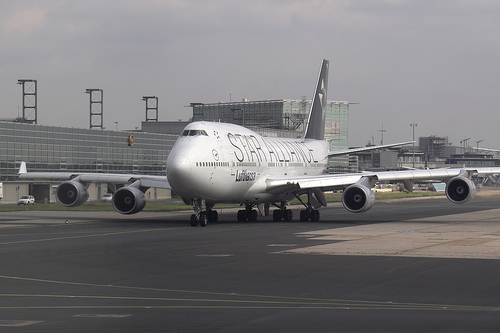}
\par\smallskip
    \includegraphics[width=47px, height=34px]{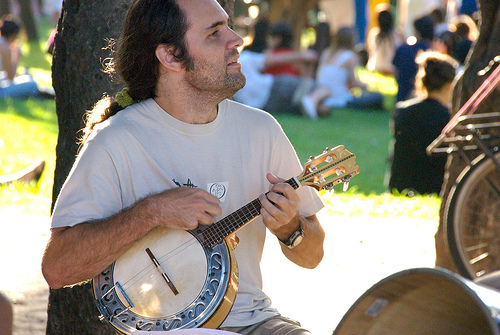}
    \includegraphics[width=47px, height=34px]{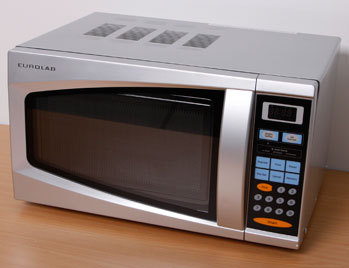}
    \includegraphics[width=47px, height=34px]{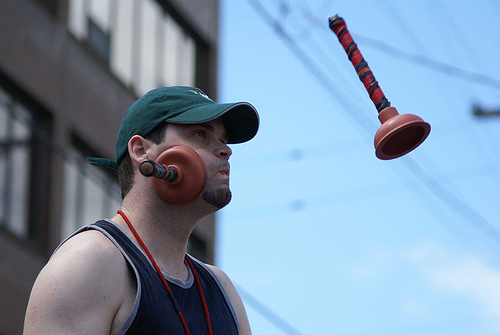}
    \includegraphics[width=47px, height=34px]{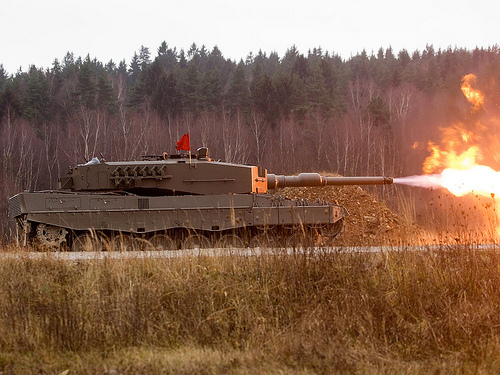}
    \includegraphics[width=47px, height=34px]{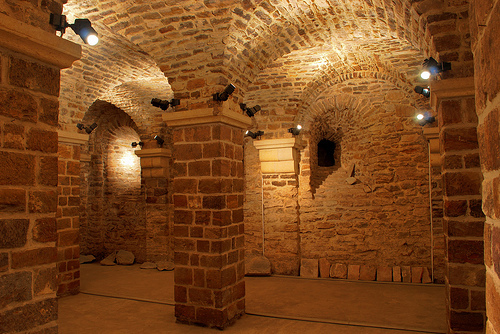}
    \caption{Sample images from the ImageNet dataset}
\end{subfigure}
 \par\medskip
\begin{subfigure}{\linewidth}
  \centering
    \fbox{\includegraphics[width=40px,height=52px]{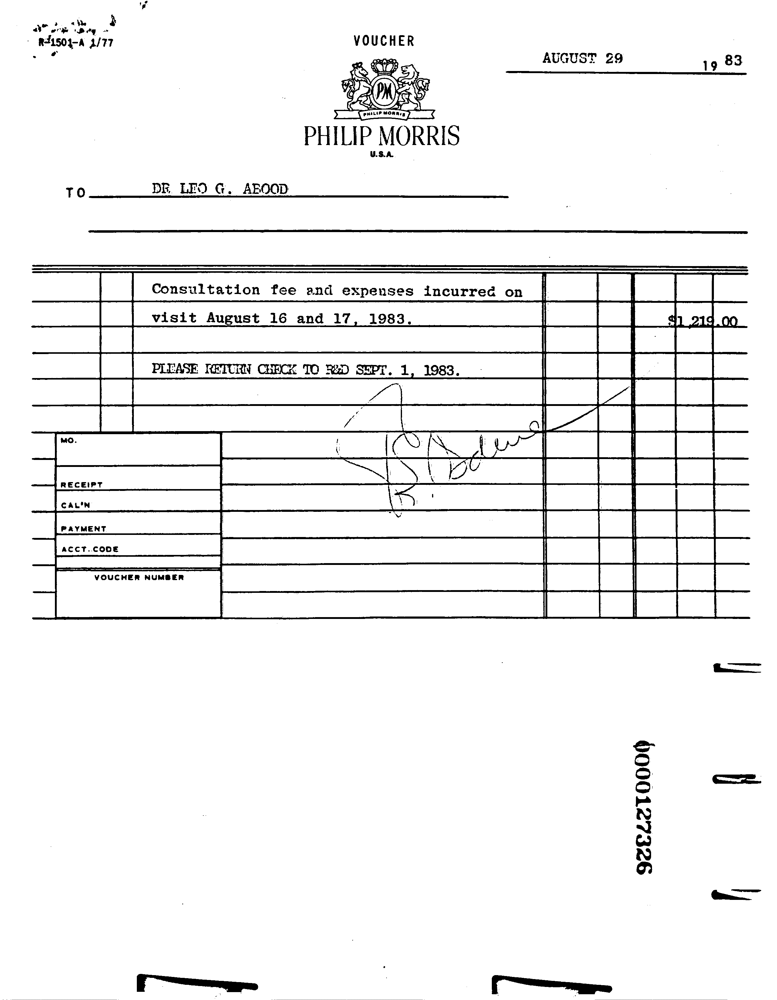}}
    \fbox{\includegraphics[width=40px,height=52px]{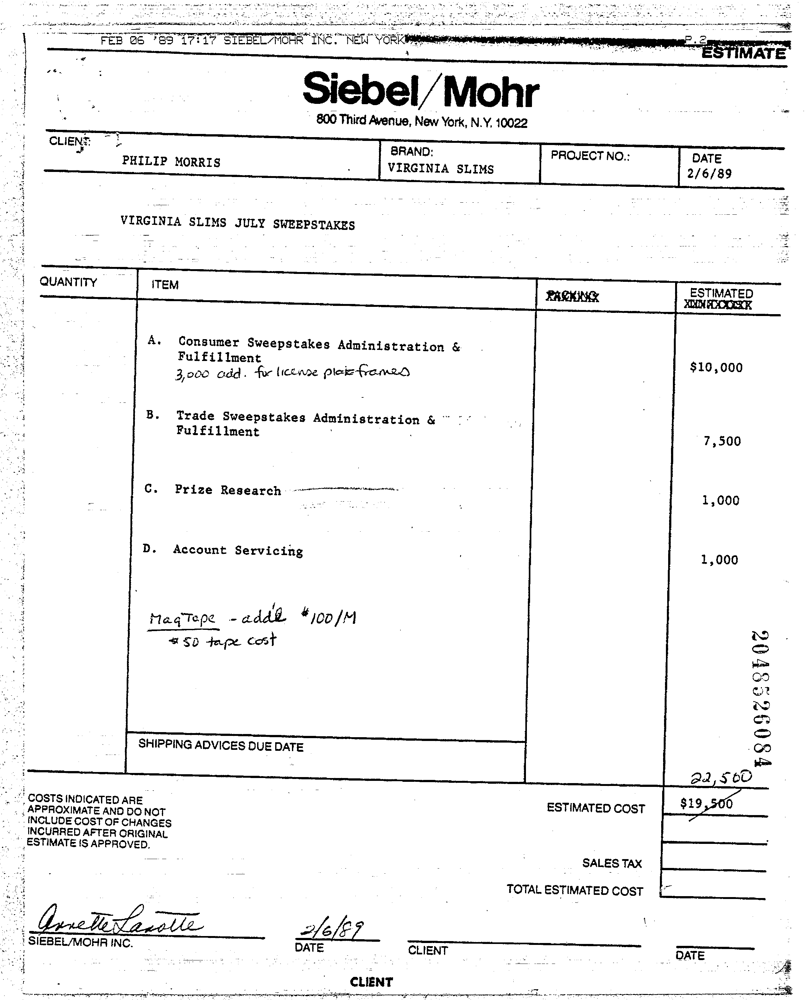}}
    \fbox{\includegraphics[width=40px,height=52px]{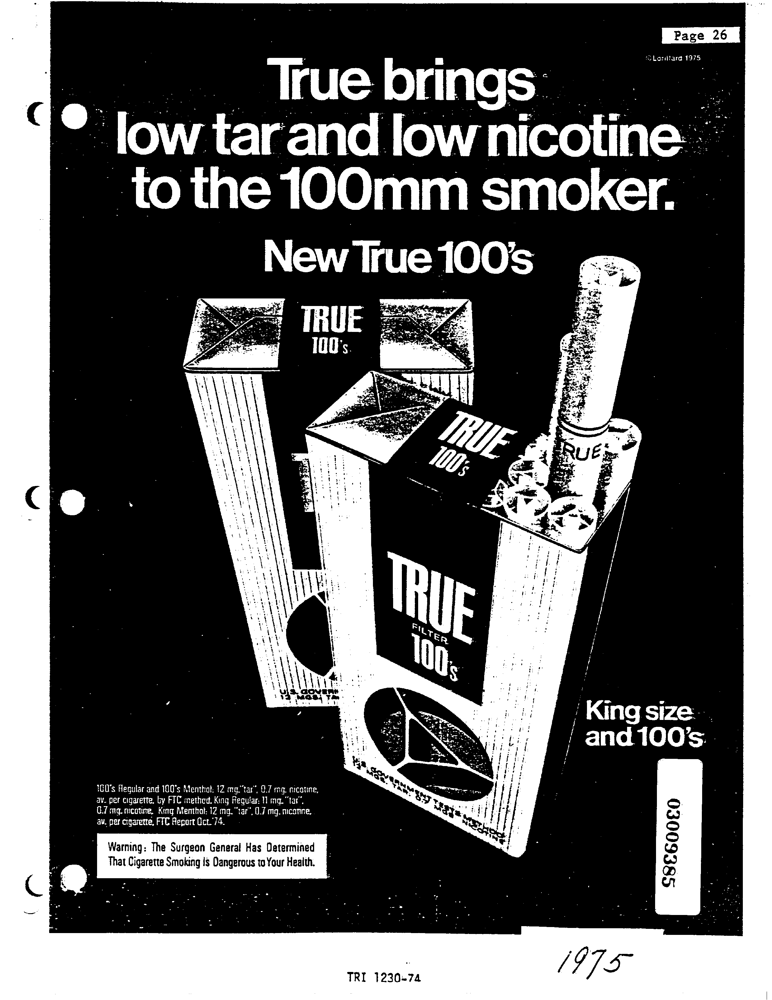}}
    \fbox{\includegraphics[width=40px,height=52px]{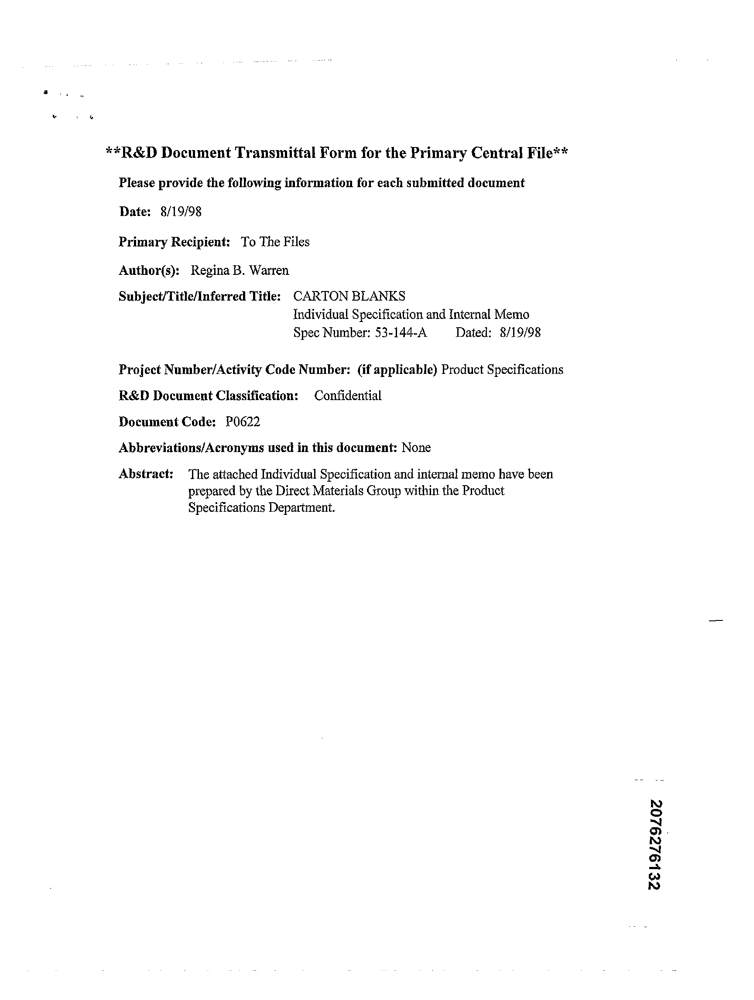}}
    \fbox{\includegraphics[width=40px,height=52px]{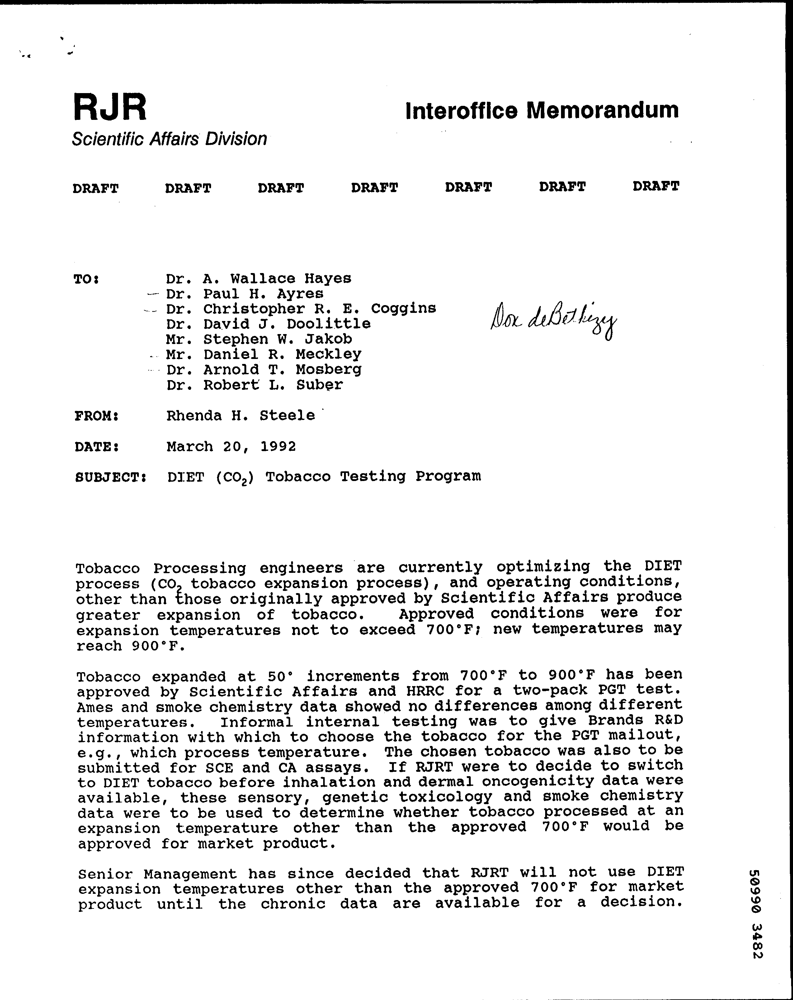}}
    \caption{Sample images from the RVL-CDIP dataset}
\end{subfigure}
\caption{Sample Images from Imagenet and RVL-CDIP datasets are shown in (a) and (b) respectively. 
}
\label{fig:imagenet}
\end{figure}

\begin{figure*}
\begin{subfigure}{.12\linewidth}
  \centering
    \fbox{\includegraphics[width=.88\linewidth, height=70px]{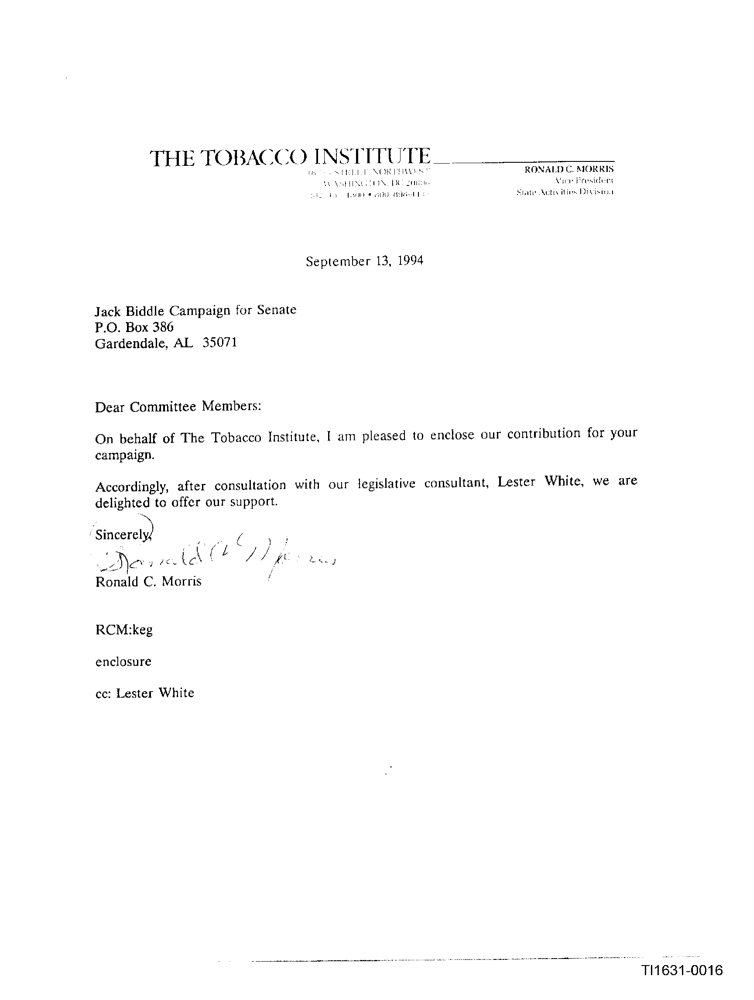}}
\end{subfigure}
\begin{subfigure}{.12\linewidth}
  \centering
    \fbox{\includegraphics[width=.88\linewidth, height=70px]{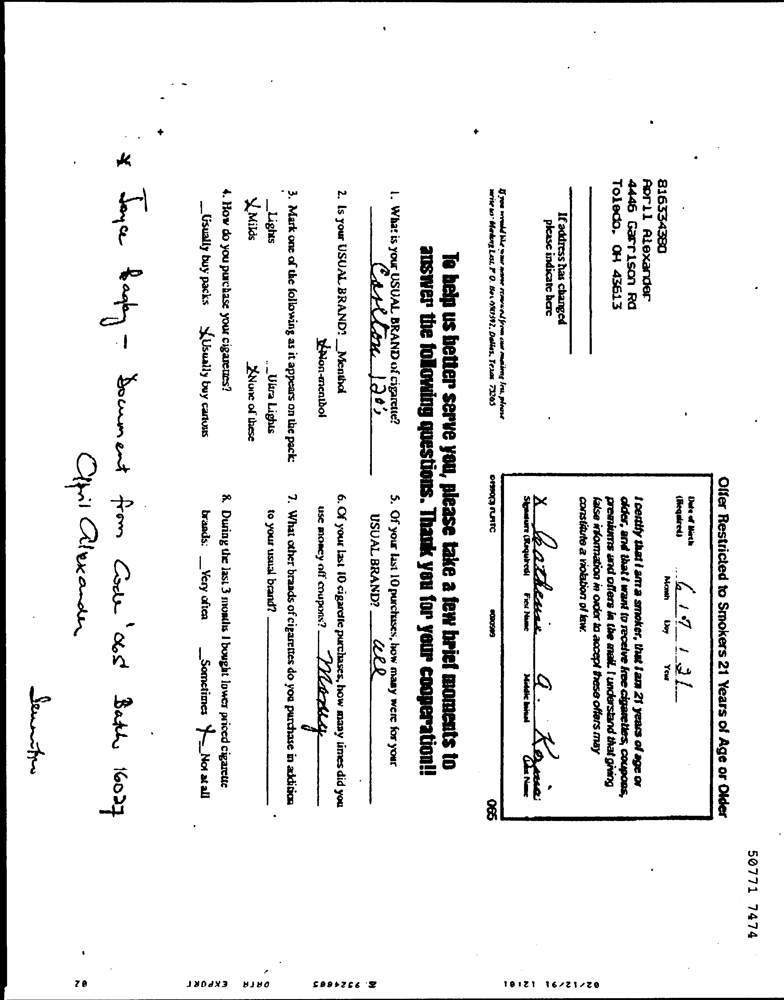}}
\end{subfigure}
\begin{subfigure}{.12\linewidth}
  \centering
    \fbox{\includegraphics[width=.89\linewidth, height=70px]{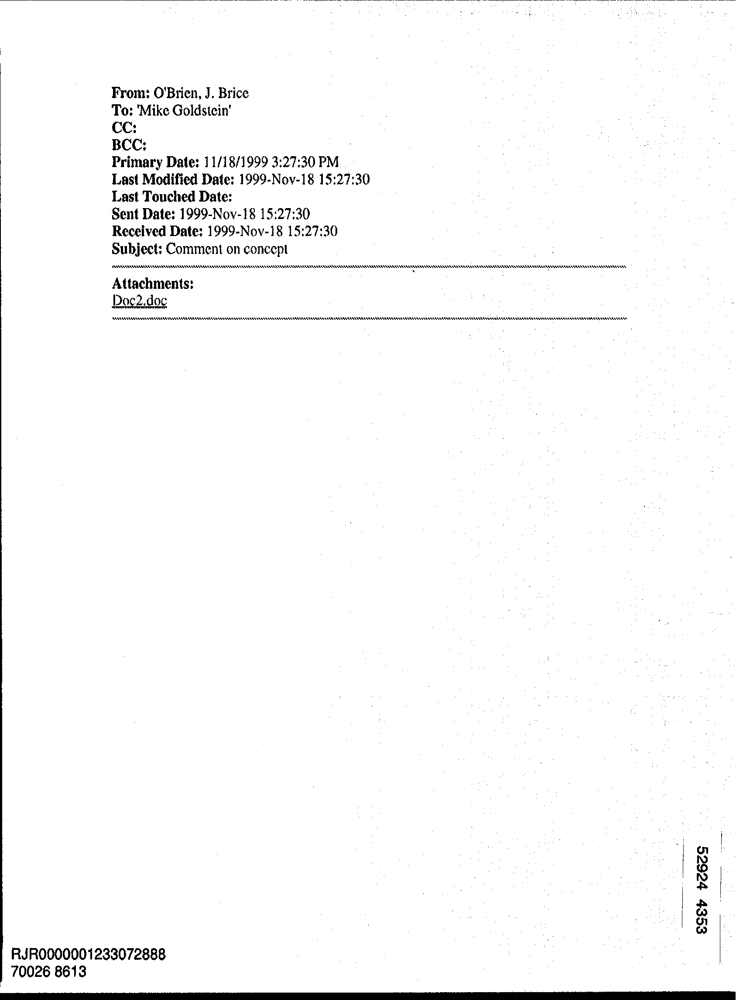}}
\end{subfigure}
\begin{subfigure}{.12\linewidth}
  \centering
    \fbox{\includegraphics[width=.89\linewidth, height=70px]{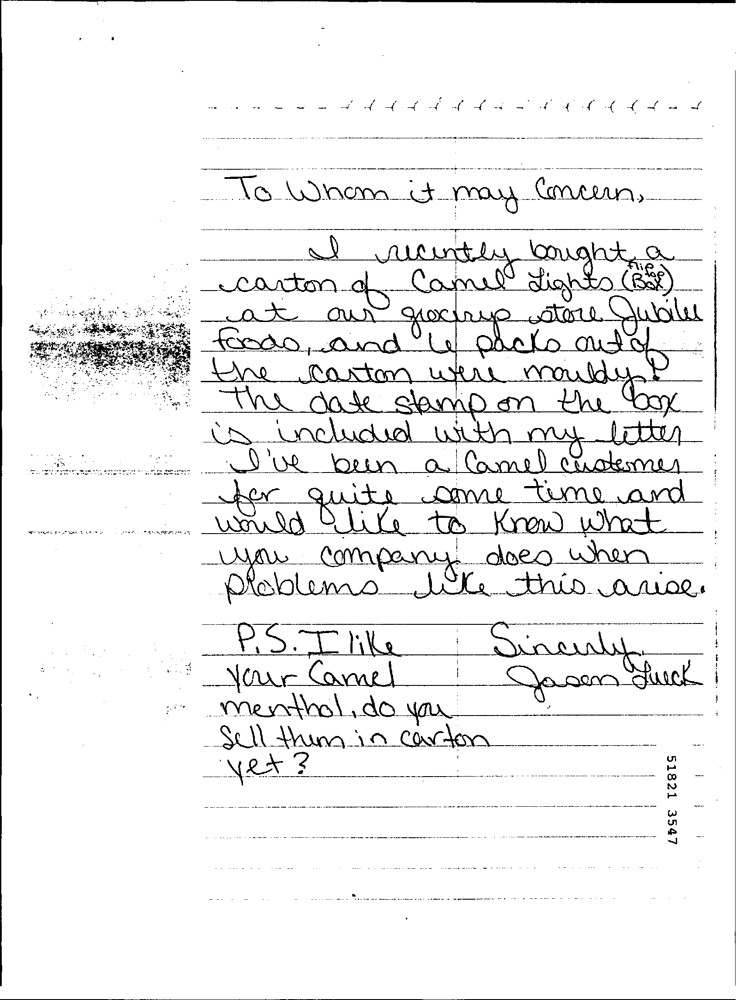}}
\end{subfigure}
\begin{subfigure}{.12\linewidth}
  \centering
    \fbox{\includegraphics[width=.89\linewidth, height=70px]{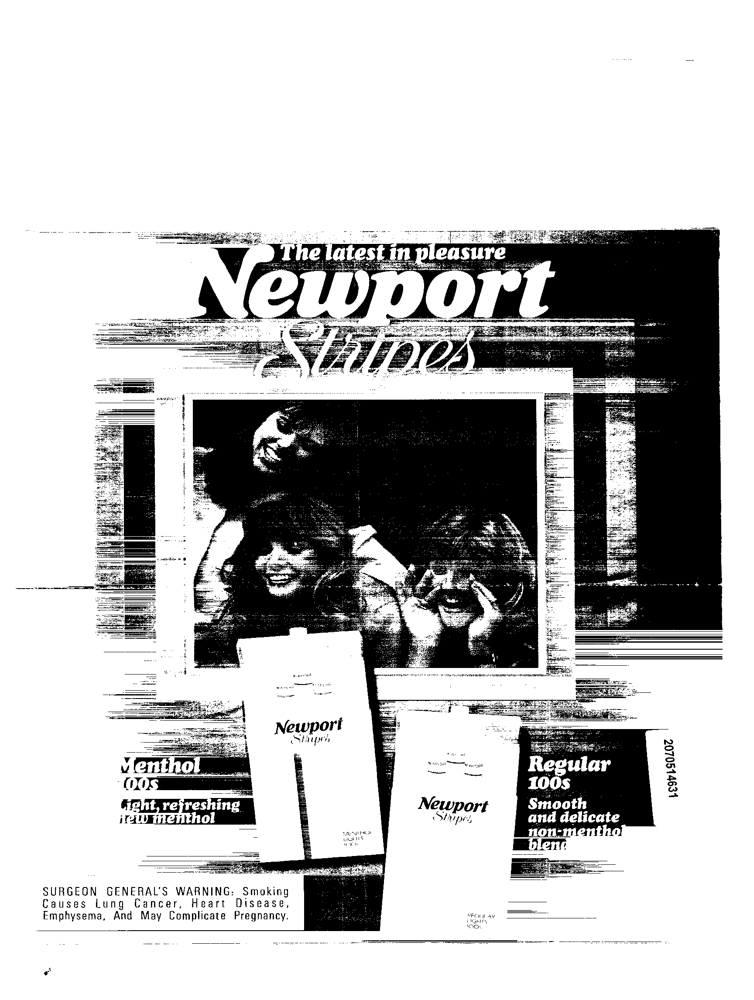}}
\end{subfigure}
\begin{subfigure}{.12\linewidth}
  \centering
    \fbox{\includegraphics[width=.89\linewidth, height=70px]{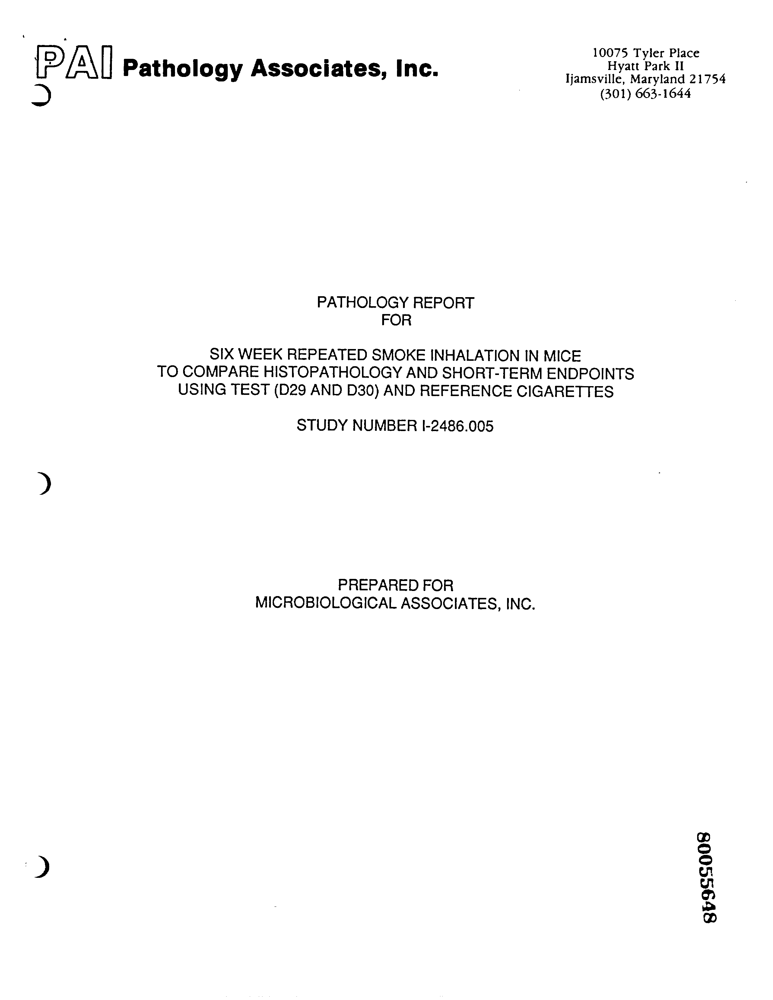}}
\end{subfigure}
\begin{subfigure}{.12\linewidth}
  \centering
    \fbox{\includegraphics[width=.89\linewidth, height=70px]{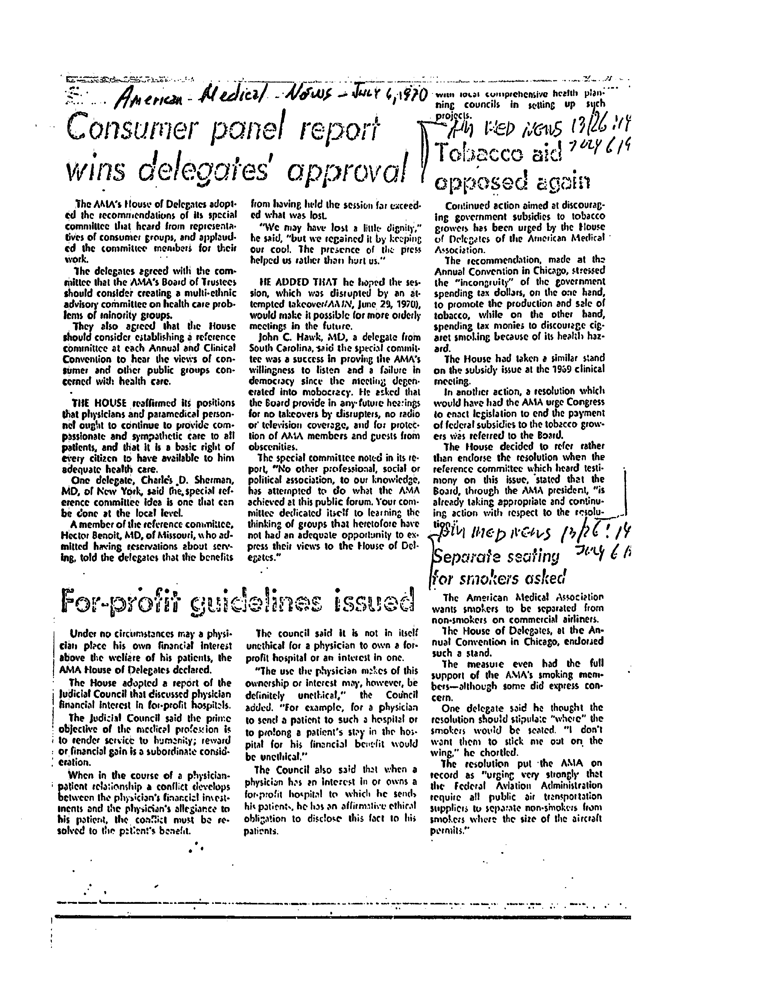}}
\end{subfigure}
\begin{subfigure}{.12\linewidth}
  \centering
    \fbox{\includegraphics[width=.89\linewidth, height=70px]{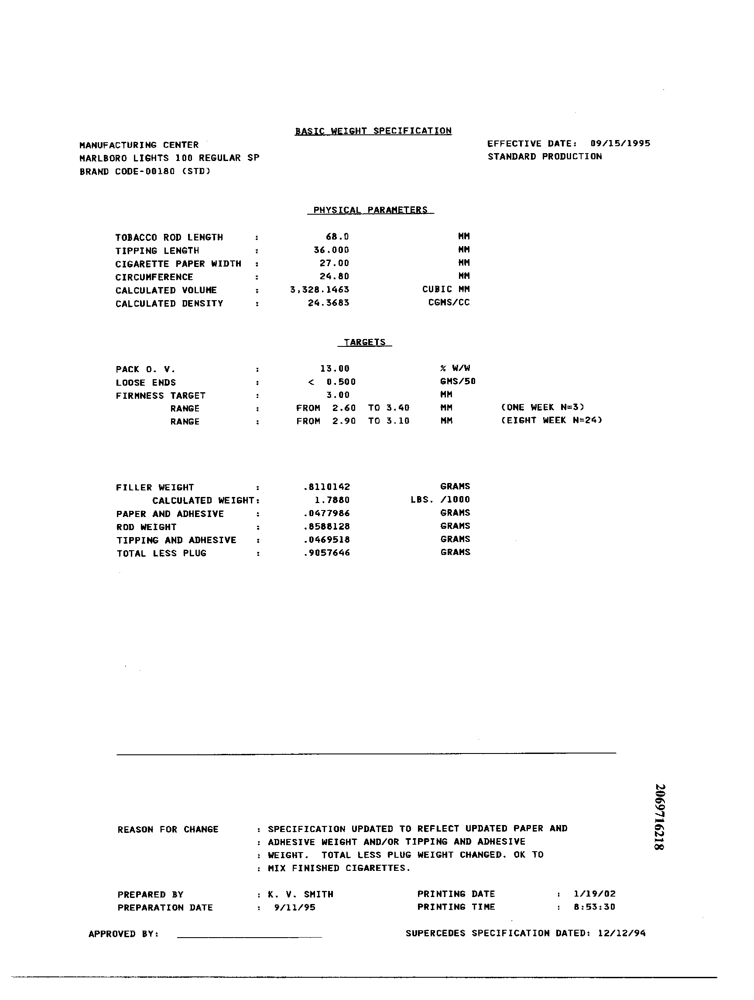}}
\end{subfigure}
\par\smallskip
\begin{subfigure}{.12\linewidth}
  \centering
    \fbox{\includegraphics[width=.88\linewidth, height=70px]{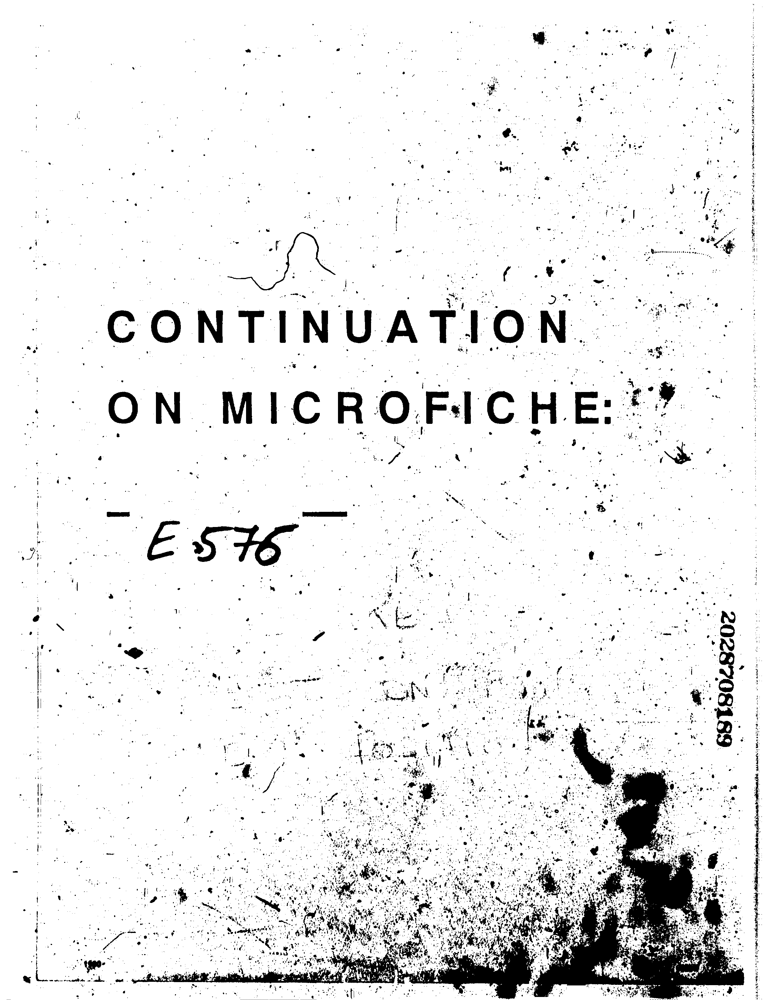}}
\end{subfigure}
\begin{subfigure}{.12\linewidth}
  \centering
    \fbox{\includegraphics[width=.89\linewidth, height=70px]{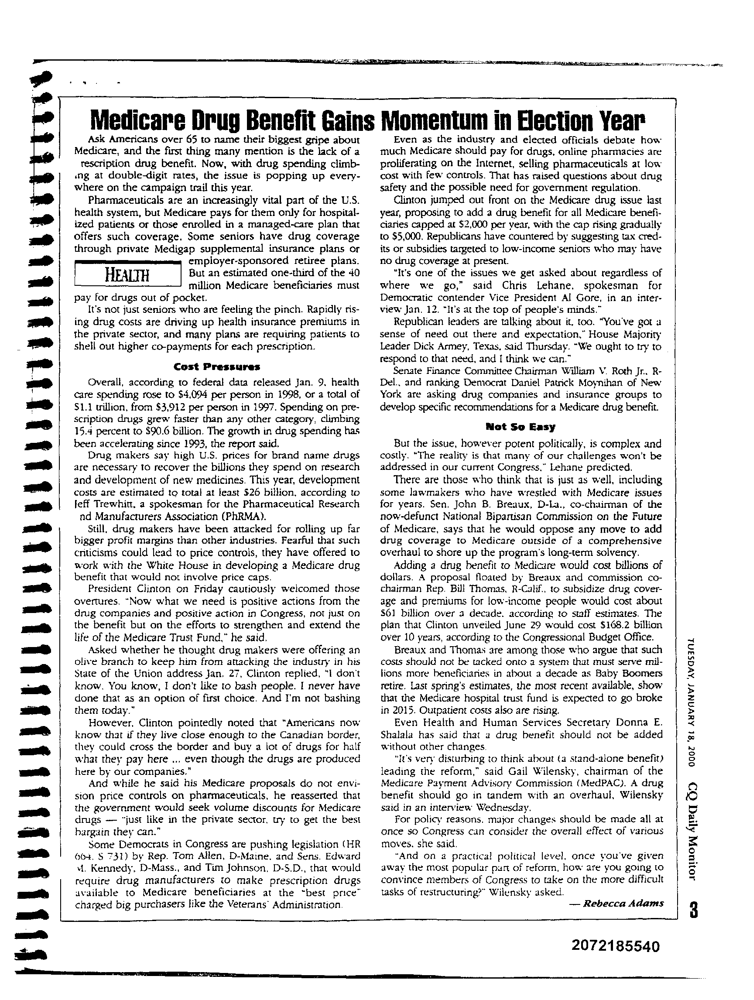}}
\end{subfigure}
\begin{subfigure}{.12\linewidth}
  \centering
    \fbox{\includegraphics[width=.89\linewidth, height=70px]{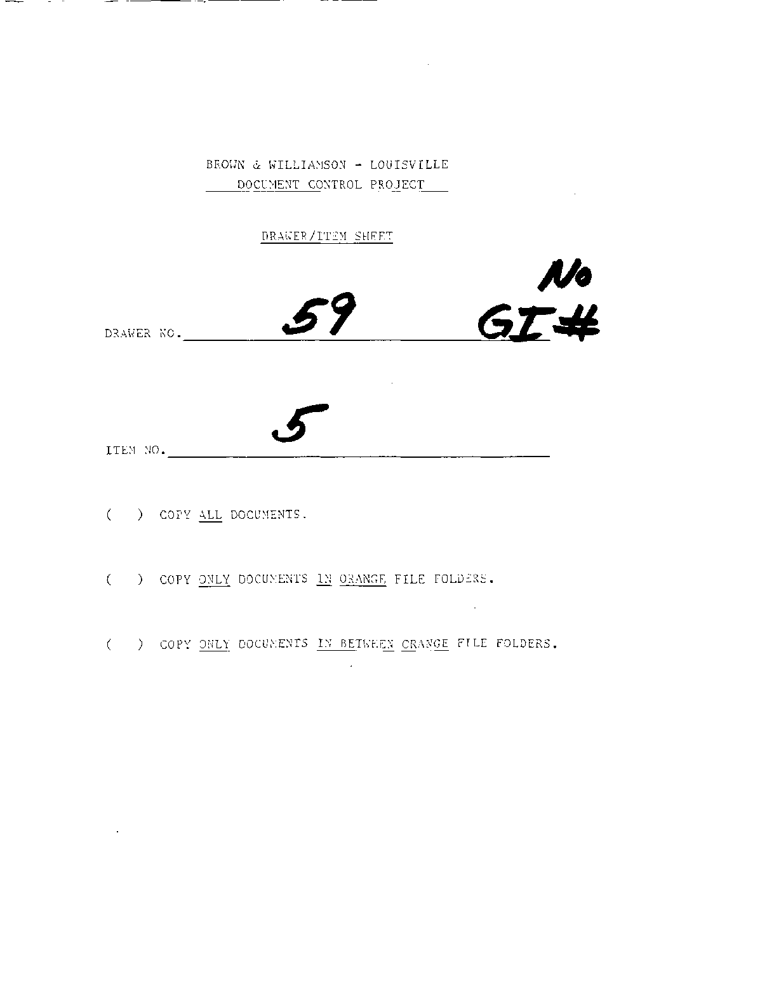}}
\end{subfigure}
\begin{subfigure}{.12\linewidth}
  \centering
    \fbox{\includegraphics[width=.89\linewidth, height=70px]{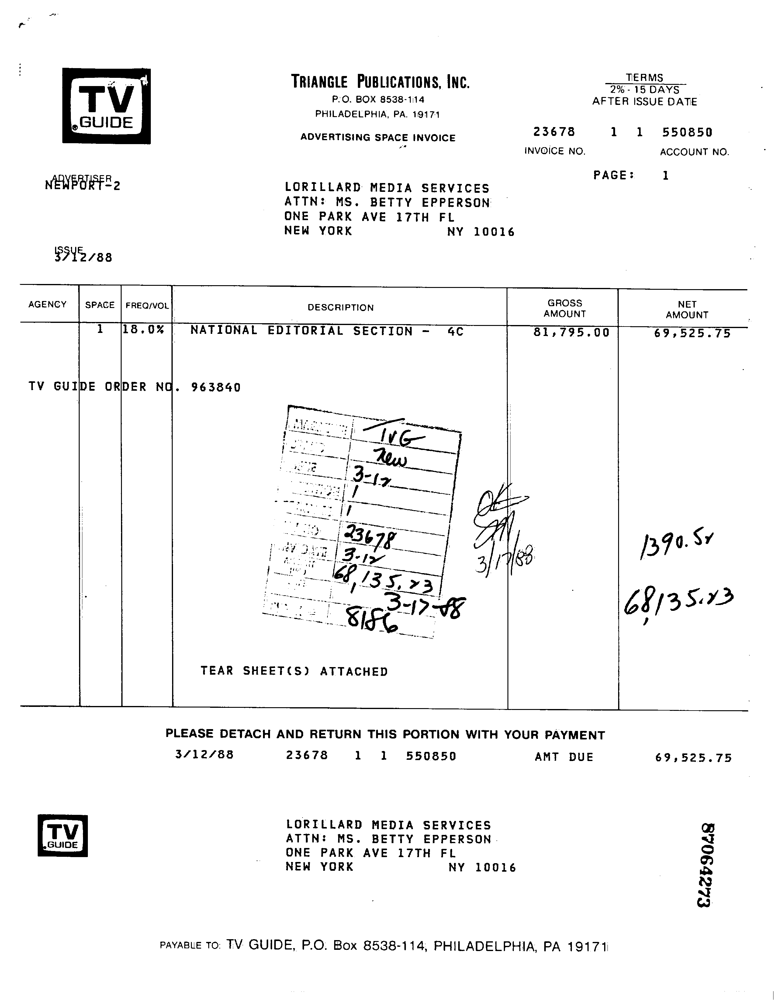}}
\end{subfigure}
\begin{subfigure}{.12\linewidth}
  \centering
    \fbox{\includegraphics[width=.89\linewidth, height=70px]{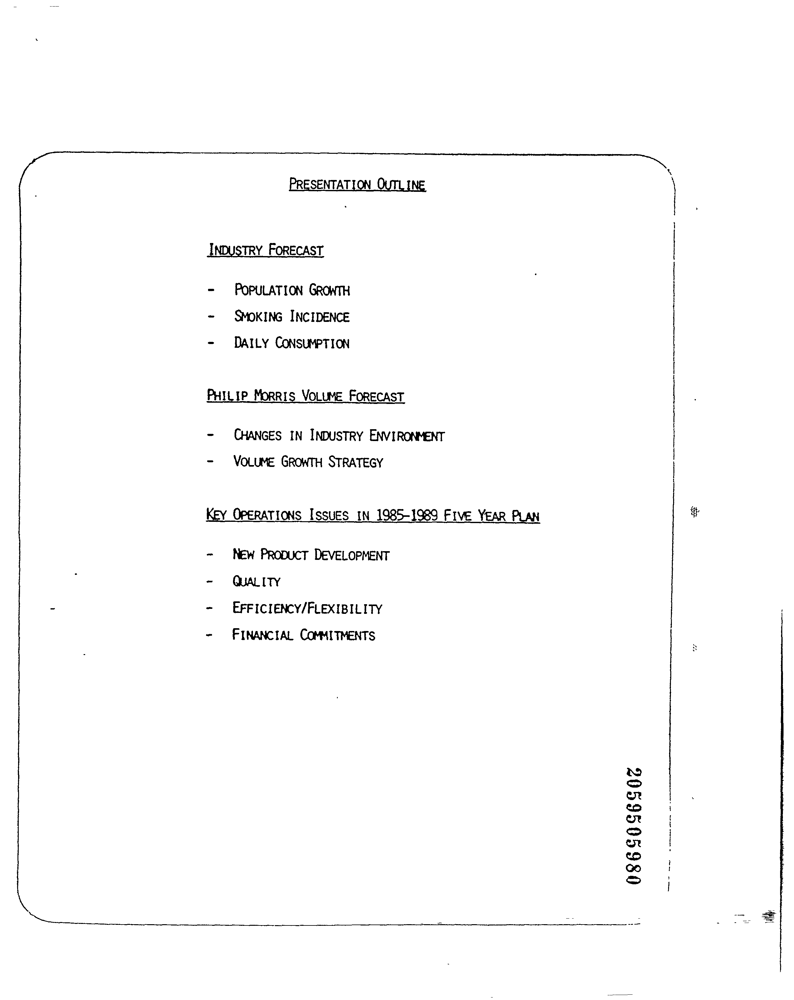}}
\end{subfigure}
\begin{subfigure}{.12\linewidth}
  \centering
    \fbox{\includegraphics[width=.89\linewidth, height=70px]{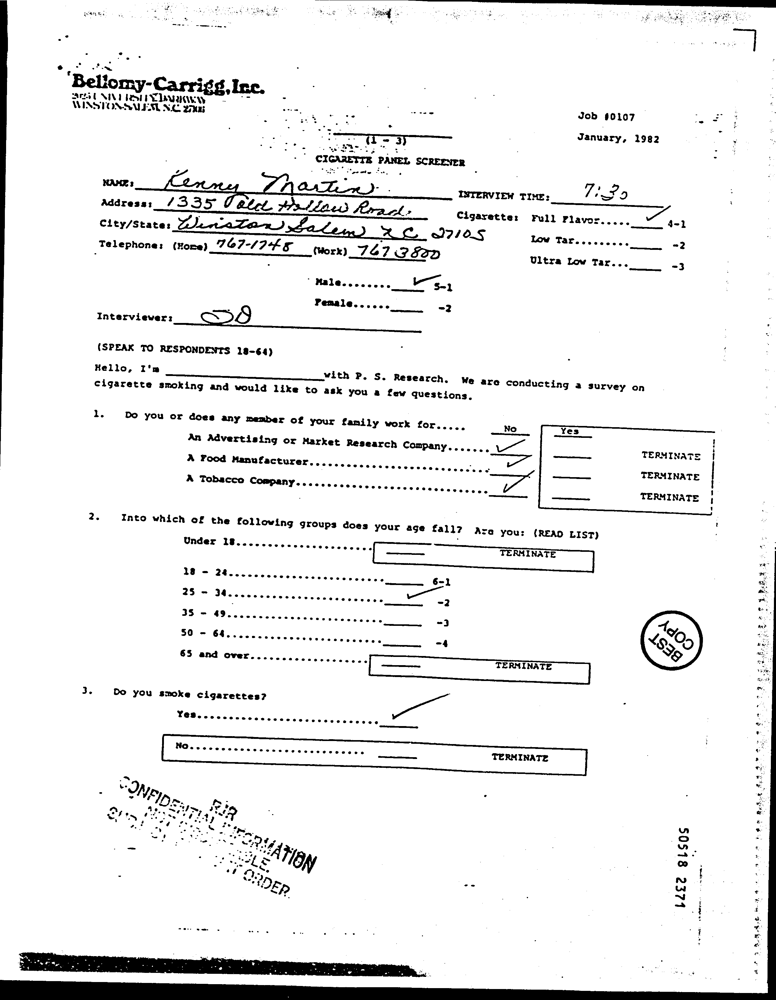}}
\end{subfigure}
\begin{subfigure}{.12\linewidth}
  \centering
    \fbox{\includegraphics[width=.89\linewidth, height=70px]{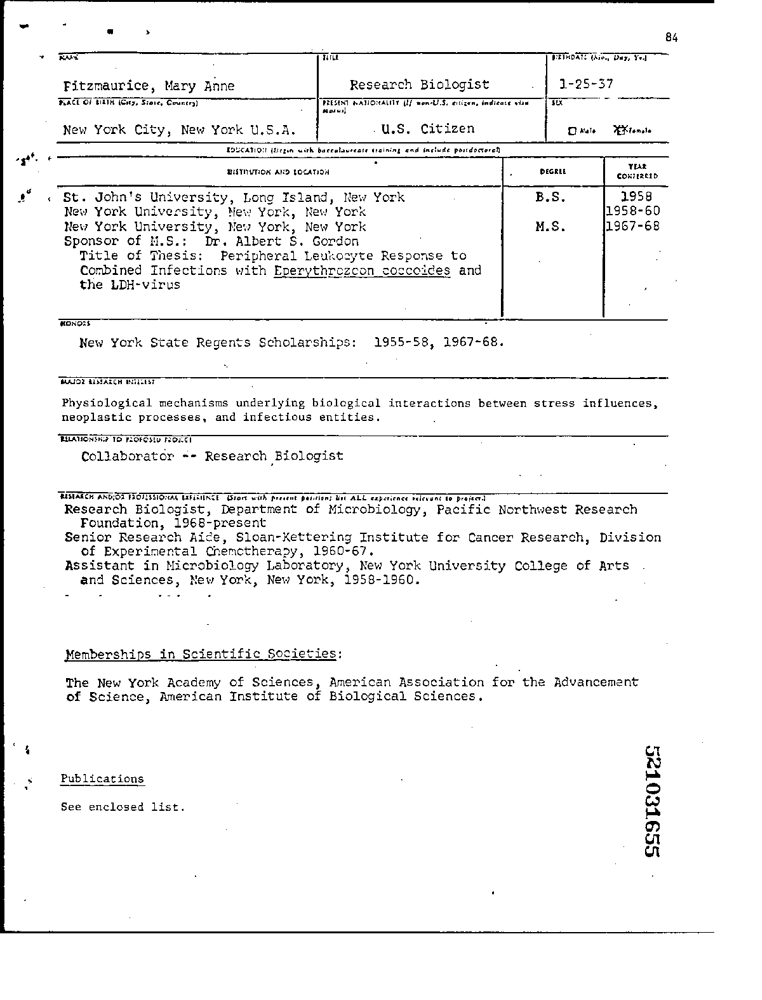}}
\end{subfigure}
\begin{subfigure}{.12\linewidth}
  \centering
    \fbox{\includegraphics[width=.89\linewidth, height=70px]{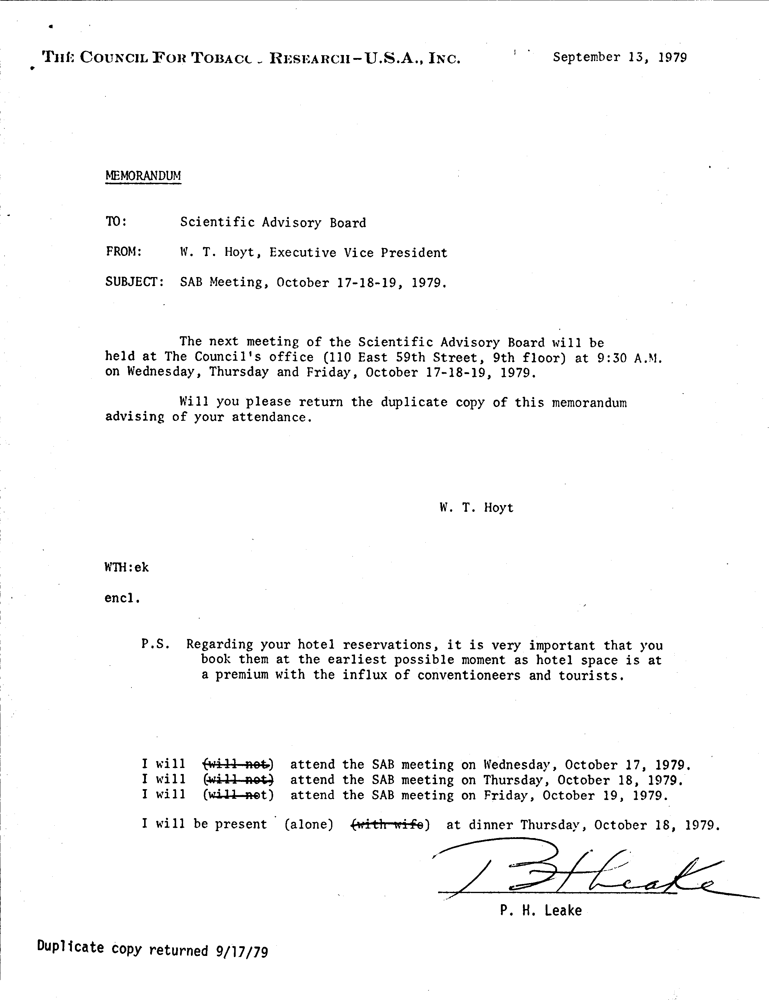}}
\end{subfigure}
\caption{Sample images from the RVL-CDIP dataset. One image from each class is depicted. From left to right: \emph{Letter}, \emph{Form}, \emph{Email}, \emph{Handwritten}, \emph{Advertisement}, \emph{Scientific report}, \emph{Scientific publication}, \emph{Specification}, \emph{File folder}, \emph{News article}, \emph{Budget}, \emph{Invoice}, \emph{Presentation}, \emph{Questionnaire}, \emph{Resume}, \emph{Memo}}
\label{fig:docimages}
\end{figure*}

Deep Learning has been used for many document analysis tasks such as binarization~\cite{afzal2015documentbin, pastor2015insights}, layout analysis~\cite{pastor2016complete, seuret2017pca}, \ac{ocr}~\cite{liwicki2007novel, breuel2013high, ahmad2015scale, ahmed2016evaluation, ahmed2016generic, ul2015sequence} etc.
Recently, deep learning methods have also been exploited for document image classification~\cite{afzal2015deepdocclassifier, lekang_14_a, harley2015evaluation}.
Deep learning methods do not require any manual feature extraction.
However, the existing \sota methods do transfer learning. 
Fig.~\ref{fig:imagenet} shows the sample images both real and document images from the ImageNet~\cite{russakovsky2015imagenet} and Tobacco-3482 datasets respectively. 
While the images are visually very different, the visual queues are generic and thus, transfer learning helps to boost the performance of the document image classification~\cite{afzal2015deepdocclassifier, harley2015evaluation}.
The networks that are not using transfer learning (i.e., they are randomly initialized) are under-performing~\cite{lekang_14_a}.
The performance evaluation for the deep neural networks was only performed using Tobacco-3482 images. 
Another dataset introduced by Harley et al.~\cite{harley2015evaluation} consist of $400,0000$ images that are divided into $16$ classes. Representative images from each of the classes are shown in Fig.~\ref{fig:docimages}.
Although now we have a large dataset available for training document images, there is no study that shows the performance of very deep networks for large datasets of document images. Furthermore, the potential of pretraining using document images only is not explored either.

Therefore, in this work, we evaluate deep neural networks both on the small and big dataset.
An exhaustive evaluation of the deep neural networks is performed to show the impact of the amount of data for training in combination with very deep \ac{cnn}. Furthermore, an evaluation is performed for transfer learning when the weights are adapted both from natural images and document images. The proposed approach shows a significant improvement over the current \sota by reducing the error by more than half.

\section{Related Work}


Over the years, different methods have been proposed for document image classification. The overall classification methods can be divided into three distinct categories.
The first category exploits structure and layout similarities, while the second focuses on developing local and global features that could be used for document classification. The third category is based on deep \ac{cnn}s that extract the features automatically for document classification. This section provides a summary of the important related work regarding the above mentioned three categories.

\begin{figure*}
    \begin{subfigure}{0.22\linewidth}
        \centering
        \fbox{\includegraphics[height=0.65\textheight]{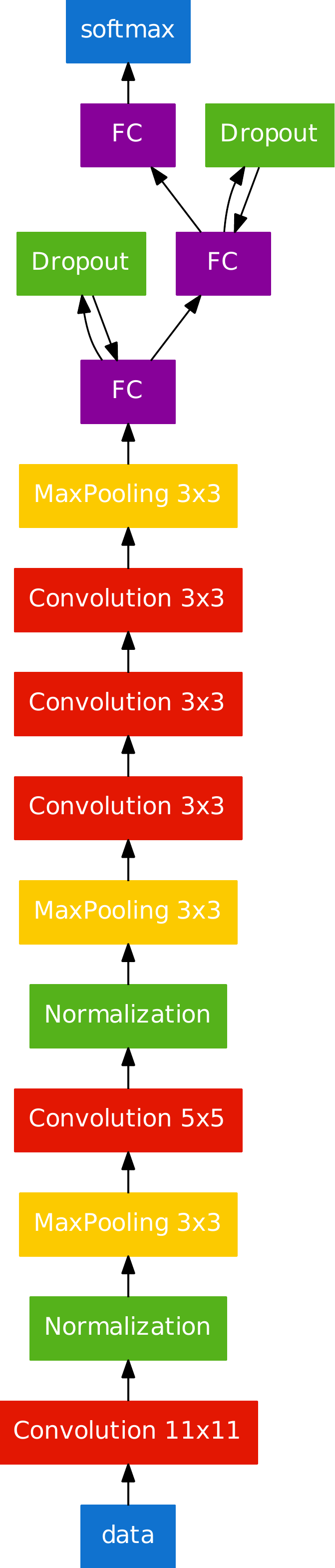}}
        \subcaption{AlexNet}
        \label{fig:alexnet}
    \end{subfigure}
    \begin{subfigure}{0.21\linewidth}
        \centering
        \fbox{\includegraphics[height=0.65\textheight]{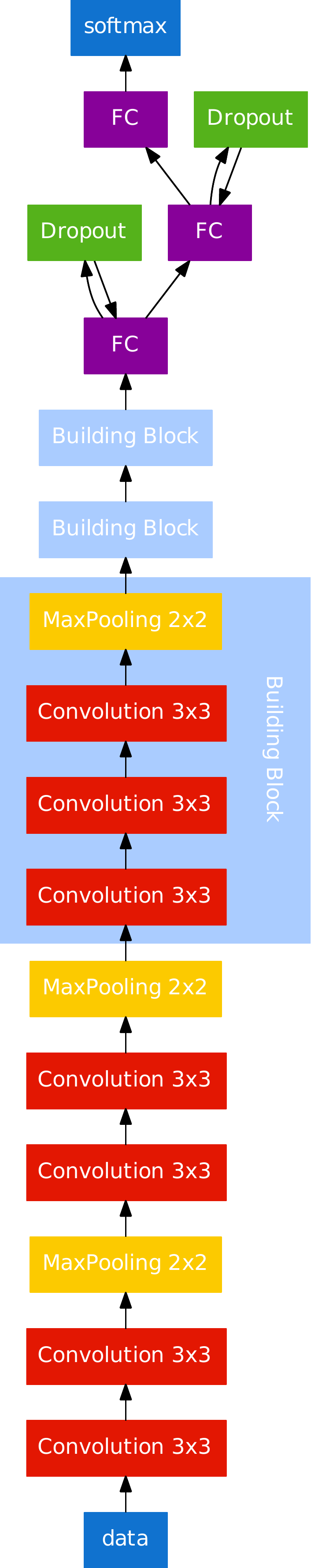}}
        \subcaption{VGG-16}
        \label{fig:vgg}
    \end{subfigure}
    \begin{subfigure}{0.36\linewidth}
        \centering
        \fbox{\includegraphics[height=0.65\textheight]{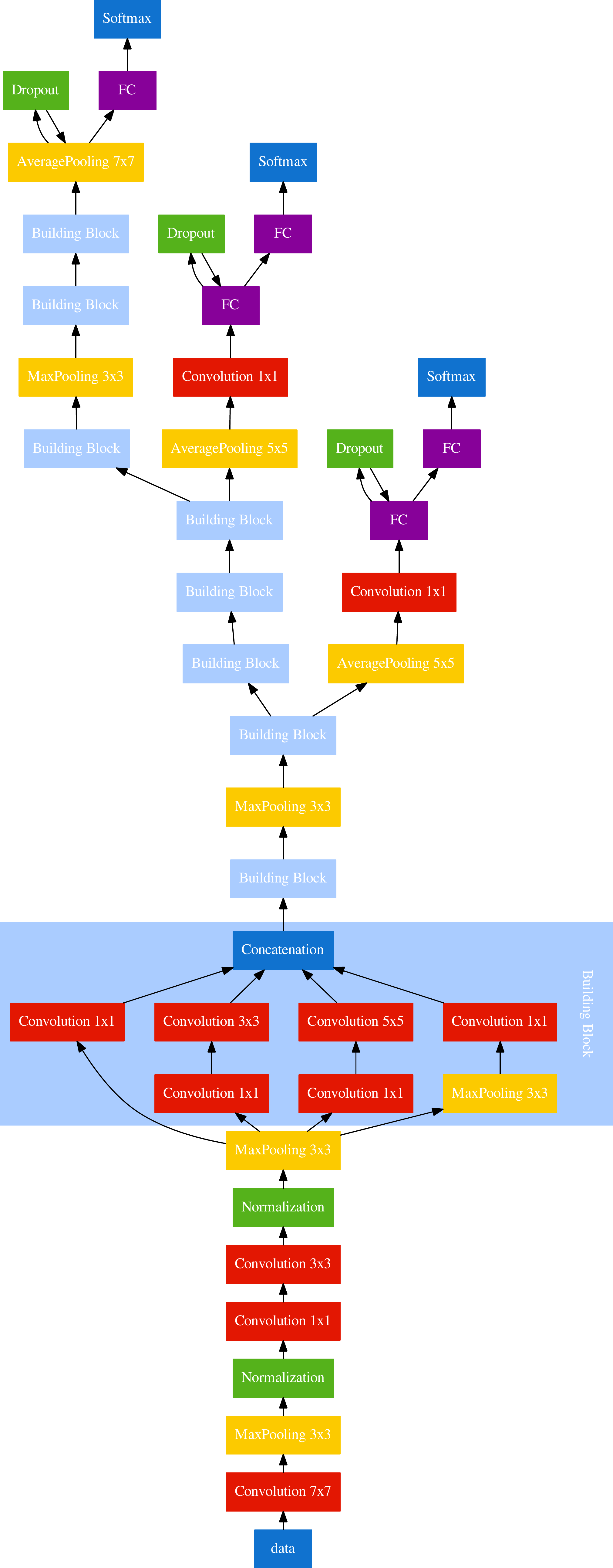}}
        \subcaption{GoogLeNet}
        \label{fig:googlenet}
    \end{subfigure}
    \begin{subfigure}{0.19\linewidth}
        \centering
        \fbox{\includegraphics[height=0.65\textheight]{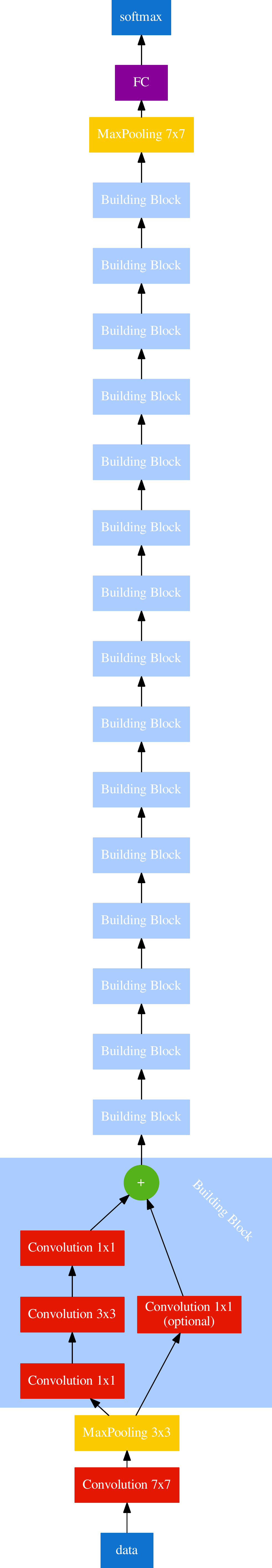}}
        \subcaption{ResNet-50}
        \label{fig:resnet}
    \end{subfigure}
    \caption{Deep CNN architectures used in this work}
    \label{fig:deepcnn}
\end{figure*}

Dengel and Dubiel~\cite{doclass_Dengel95} used layout structure printed documents. They used top-down induction in decision trees to convert printed documents into a complementary logical structure.
Bagdanov and Worring~\cite{doclass_Bagdanov2001} classify machine-printed documents by using the Attributed Relational Graphs (ARGs).
Byun and Lee~\cite{doclass_Byun2000} used parts of the documents for the recognition. They reasoned that processing complete documents is time-consuming. The document classification was performed on parts of the documents using DP algorithm. Their approach was fast but the applicability is limited to forms. Shin and Doermann~\cite{doclass_shin} proposed an approach that used layout structural similarity for full or partial image matching for retrieval. 
Kevyn and Nickolov~\cite{Collins-thompson02aclustering-based} used both the layout and the text features for matching the documents for retrieval. 

Jayant et al.~\cite{doclass_Kumar12} propose a method that relies on the patch code words derived from the document images. The code book is learned independently of the class labels of the documents. In the first step, the images are recursively partitioned both in horizontal and vertical direction for modeling spatial relationships. Subsequently, a histogram for each partition is calculated that is used for the classification.
Following the same idea of developing the code book, another work presented by Jayant et al.~\cite{doclass_Kumar14} build a codebook of SURF descriptors extracted from training images. Then, histograms of codewords are created similar to~\cite{doclass_Kumar12}. A Random Forest classifier is used for classification. The applicability of the approach is shown in the presence of limited data.
Chen et al.~\cite{doclass_Chen12} propose a method based on low-level image features to classify documents. The approach is limited to structured documents. An important point is that one could obtain the registration of two images by matching the feature points.
Joutel et al.~\cite{doclass_Joutel2007} proposed a method that used curvelet transformation for indexing and querying the documents at different image scales. Their method is designed particularly for large databases of handwritten manuscripts. Kochi and Saitoh~\cite{doclass_Kochi99} used textual descriptions of document images for information extraction from documents. The method is limited to semi-structured documents and assumes a pre-defined knowledge is available for the document classes.
Reddy and Govindaraju~\cite{doclass_umamaheswara08} used binary images for the classification of the documents. They use pixel information and calculate pixel densities.  They used k-means clustering supported by adaptive boosting. The method is evaluated on the benchmark NIST scanned special tax form databases $2$ and $6$.

The pioneering work that performed document classification using \ac{cnn}s used a rather shallow network for classification~\cite{lekang_14_a}. Nevertheless, the proposed approach outperformed structural similarity based methods and shows the potential of automatic feature learning for document classification using \ac{cnn}s. The reason may be that deep networks require a lot of data for training and at that time the standard challenging dataset consisted of only $3,482$ images.
Afzal et. al.~\cite{afzal2015deepdocclassifier} and Harley et. al.~\cite{harley2015evaluation} provided a breakthrough when they showed that it is possible to use transfer learning and the features that are learned from general (daily life) images can be used for the classification of document images~\cite{afzal2015deepdocclassifier}. They achieved a significant improvement over \ac{cnn} based methods that were the \sota at that time.
Another notable contribution by Harley et. al.~\cite{harley2015evaluation} was that they introduced a dataset consisting of $400,000$ documents divided into $16$ classes.
This allowed for the evaluation of deep neural networks using a significant amount of data. 

The \sota in deep \ac{cnn}s has advanced significantly in recent years and there has been no comprehensive study regarding the impact of deep architectures for document classification. Moreover, there is no study that explores transfer learning from document images and also there is no report of the impact of the amount of training images. The presented work takes into account these issues and performs a comprehensive set of experiments to fill the gaps that exist. Eventually, this study leads to an approach that can reduce the error by more than half and therefore provides another leap forward in the domain of document image classification.


\section{Deep Convolutional Neural Networks}
\label{sec:networks}

This section briefly presents the deep \ac{cnn} architectures used in this work. Furthermore, the image preprocessing and training details are described.

\subsection{Network Architectures}

The deep \ac{cnn} architectures used in this paper are well known in the domain of object recognition but are not used frequently for document image classification. The networks are of very different nature (\cf~Fig.~\ref{fig:deepcnn}).

\subsubsection{\textbf{AlexNet}}
AlexNet \cite{cnn_alexnet_nips2014} is the eight-layer \ac{cnn} that won the ImageNet Large Scale Visual Recognition Challenge (ILSVRC) in 2012 \cite{russakovsky2015imagenet} by a large margin.
It employs five convolutional layers with optional pooling and local response normalization. These are then followed by three fully-connected layers and a softmax classifier (\cf~Fig.~\ref{fig:alexnet}).

\subsubsection{\textbf{VGG-16}}
VGG-16, as the name suggests is a 16-layer \ac{cnn}~\cite{simonyan2014very}. Unlike AlexNet, it uses only convolutional filters of size $3\times3$. Just like AlexNet, it has a straightforward architecture, but with $13$ convolutional layers and $3$ fully connected layers (\cf~Fig.~\ref{fig:vgg}) it is quite a bit deeper and has a repetitive pattern of layers. This architecture has won the localization category of the ILSVRC 2014.

\subsubsection{\textbf{GoogLeNet}}
GoogLeNet, just like VGG-16, won a category of the ILSVRC 2014, namely the classification category~\cite{szegedy2015going}. The architecture of this network, however, is a bit more sophisticated (\cf Fig.~\ref{fig:googlenet}). Unlike AlexNet and VGG-16, it is not just a stack of Convolution layers and Pooling layers, but rather a stack of building blocks, which themselves consist of Convolution and Pooling layers. It is therefore a Network-in-Network approach~\cite{lin2013network}. Due to its high depth, the network employs three softmax classifiers during training, to enable efficient backpropagation of the error. At test time, the two auxiliary classifiers are discarded.

\subsubsection{\textbf{Resnet-50}}
ResNets are a family of very deep \ac{cnn} architectures which make use of residual connections~\cite{he2016deep} to overcome the challenge of efficient error backpropagation. ResNet-50 is a variant of the network with $50$ layers, which, as in GoogLeNet, are grouped in building blocks (\cf~Fig.~\ref{fig:resnet}). An even deeper variant with $152$ layers won the ILSVRC classification task in 2015. Interestingly, despite its increased depth, the network has fewer parameters to fit than VGG-16.

\subsection{Preprocessing}

As the networks used in this paper require images of a fixed size as input, we first downscale all images to the expected input size of the networks. For AlexNet, the images are resized to $227\times227$ pixels, for the other networks the images are resized to $224\times224$ pixels.
Typically, when training \ac{cnn}s, the training data is augmented by resizing the images to a larger size, \eg $256\times256$ pixels and then cropping random patches of these images in the size of the network input. This approach has shown to be effective for real-world image classification~\cite{cnn_alexnet_nips2014}. In real-world images, the objects are typically close to the center of the image and therefore always contained in the random crops. However, the most discriminative parts of document images are not always close to the center of the image but reside in the outer regions, \eg the head of a letter. Therefore, we do not enlarge our training dataset in this way but train solely with images containing the entire document.

After resizing the images, we compute the mean pixel values of the training images and subtract them from all images to center the training data.

As a last preprocessing step, we convert the grayscale images to RGB images by simply copying the pixel values of the single-channel images to three channels.

\subsection{Training Details}
We train all networks using stochastic gradient descent with a momentum of $0.9$ and a learning rate that is updated every iteration to
\begin{equation}
    lr = initial\_lr * \left( 1 - \frac{iter}{max\_iter}\right) ^ {0.5}
\end{equation}
The initial learning rate is set to a value between $0.01$ and $0.0001$ depending on the network architecture, the training dataset and the weight initialization.

The number of training epochs depends on the task and ranges between $40$ and $80$ epochs.

\section{Experiments and Results}

\begin{table*}
\renewcommand{\arraystretch}{1.3}
\centering
\caption{Performance of the networks on the Tobacco-3482 dataset with $100$ training samples per class and different weight initializations.}
\begin{tabular}{l|c|c|c}
 & Document Pretraining & ImageNet Pretraining & No Pretraining \\\hline
AlexNet & $90.04\,\%$ & $75.73\,\%$ & $62.49\,\%$ \\\hline
GoogLeNet & $88.40\,\%$ & $72.98\,\%$ & $70.28\,\%$ \\\hline
VGG-16 & $91.01\,\%$ & $77.52\,\%$ & $69.50\,\%$ \\\hline
Resnet-50 & $91.13\,\%$ & $67.93\,\%$ & $59.55\,\%$
\end{tabular}
\label{tab:accuracy_small}
\end{table*}

\subsection{Datasets}
To evaluate the performance of the deep neural networks presented in section \ref{sec:networks}, two datasets are used. First, we train a variety of networks on the Ryerson Vision Lab Complex Document Information Processing (RVL-CDIP) dataset \cite{harley2015evaluation}. This dataset consists of $400,000$ labeled document images from 16 classes. The dataset is already split into a training dataset which contains $320,000$ images and a validation and a test dataset which each contain $40,000$ images.

Secondly, we use the Tobacco-3482 dataset \cite{doclass_Kumar14} to evaluate the performance of the deep \ac{cnn}s and to investigate to which extent transfer learning from the first dataset is applicable. The Tobacco-3482 dataset contains $3,482$ images from ten document classes. 

Both datasets are quite similar and there even exists some overlap. Therefore, at the transfer learning experiments, we pretrain the networks not on the full RVL-CDIP dataset, but only on the images that are not contained in the Tobacco-3482 dataset. Thus, the networks are pretrained on only $319,784$ training images.

\subsection{Evaluation}

For the RVL-CDIP dataset, we just train the networks and report the top-1 accuracy achieved on the test set. Since the Tobacco-3482 dataset is so small, we use a slightly more sophisticated evaluation technique to get an expected accuracy and to avoid unrepresentative results due to random initialization or a specific dataset split. To come up with a robust estimate of how well the networks perform, we split the dataset such, that $10$ to $100$ images per class are used for training while the rest are for testing. The training dataset is again split with an $80/20$ ratio, so that $20\,\%$ of the training data are used for validation. For each split size, we randomly create ten dataset partitions and report the median accuracy achieved by the networks.
This is similar to the evaluation scheme that was also used by Kang et al.~\cite{lekang_14_a} and Kumar et al.~\cite{doclass_Kumar14} and allows for a fair comparison with their approaches.

\subsection{Results on Tobacco-3482}

\begin{figure}
        \centering
        \includegraphics[width=\linewidth]{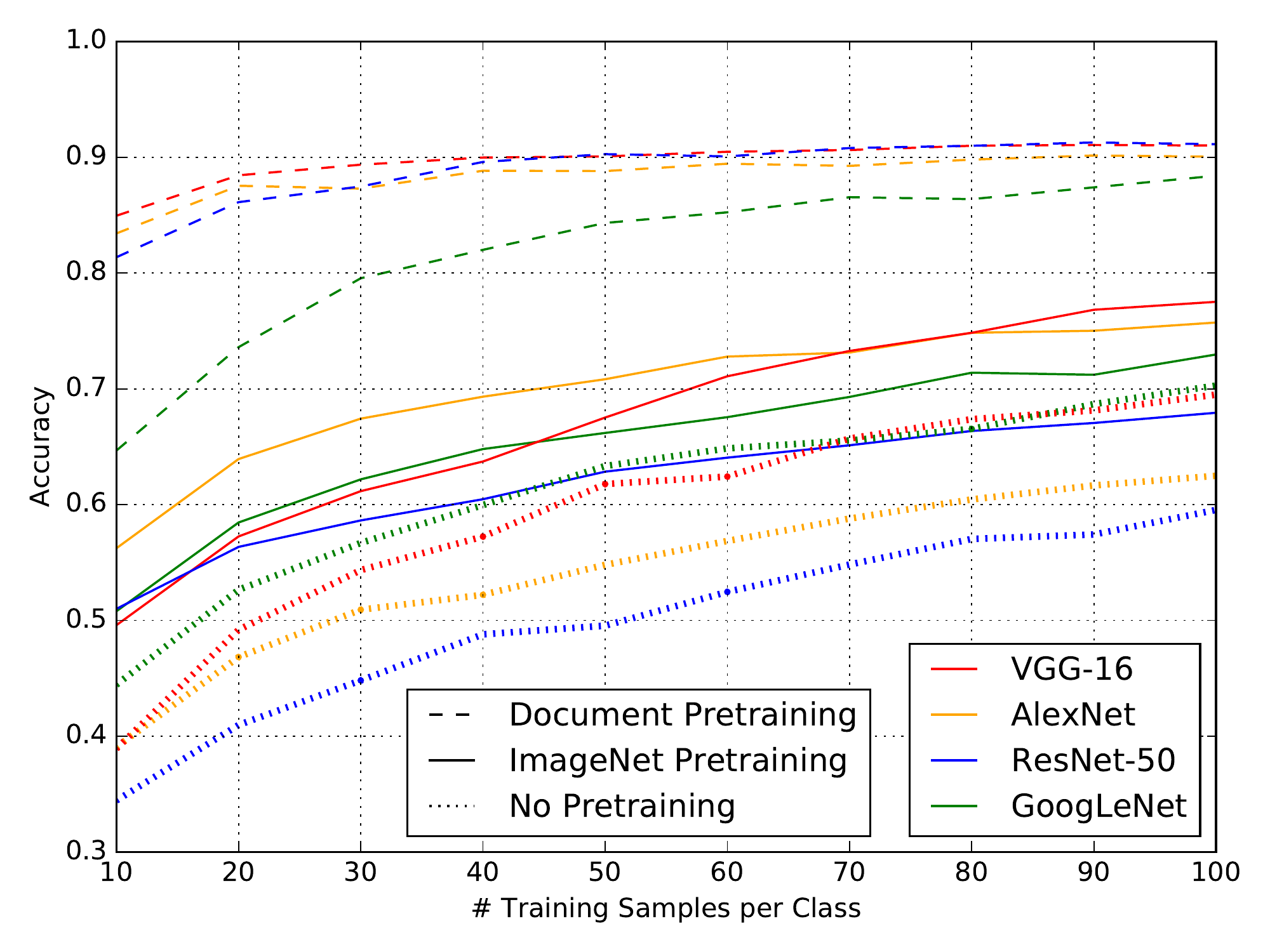}
        \caption{Mean accuracy achieved on the Tobacco-3482 dataset. The dashed graphs represent the networks that are pretrained on the RVL-CDIP dataset, the solid lines represent the network with ImageNet pretraining and the dotted lines show the network accuracy when trained from scratch.}
\label{fig:accuracy}
\end{figure}

We have trained the four deep \ac{cnn}s described in Section~\ref{sec:networks} on the two datasets with different weight initializations to investigate the benefits of transfer learning. As shown in Table~\ref{tab:accuracy_small} and Fig.~\ref{fig:accuracy} which correspond to the achieved performance on the Tobacco-3482 dataset, transfer learning does improve the classification performance significantly. When the networks are pretrained on a similar dataset, the accuracy achieved on the final dataset is higher than $90\,\%$ and already with as little training data as $10$ samples per class, we could outperform the current \sota which achieves only $77.6\,\%$~\cite{afzal2015deepdocclassifier}.

\subsection{Analysis}

We also compare the networks with ImageNet initialization against randomly initialized networks and find that even though the images are substantially different (\cf~Fig.~\ref{fig:imagenet}), it helps to use pretrained models. Fortunately, there are models which are pretrained on ImageNet available online for many architectures, including the four networks used in this work. So, in case there is no large document dataset available for pretraining, one can and should always resort to using an ImageNet pretrained model for finetuning.

Depending on the amount of available training data, AlexNet and VGG-16 are the best choices when finetuning the networks from models that were pretrained on ImageNet (\cf~Fig.~\ref{fig:accuracy}). When pretrained on the RVL-CDIP dataset, GoogLeNet is significantly worse than the other networks, especially for a small amount of training data.

\subsection{Results on RVL-CDIP}

On the large-scale RVL-CDIP dataset, all networks achieve very good results (\cf~Table~\ref{tab:accuracy_large}) with the VGG-16 performing best at an accuracy of $90.97\,\%$. The current \sota on this dataset only achieves an accuracy of $89.8\,\%$, thus we could decrease the relative error by more than $11\,\%$ by simply using a different network architecture.
Note, that even though the training dataset is quite large, all of the networks still benefit from Imagenet pretraining.

\begin{figure}
        \centering
        \includegraphics[width=\linewidth]{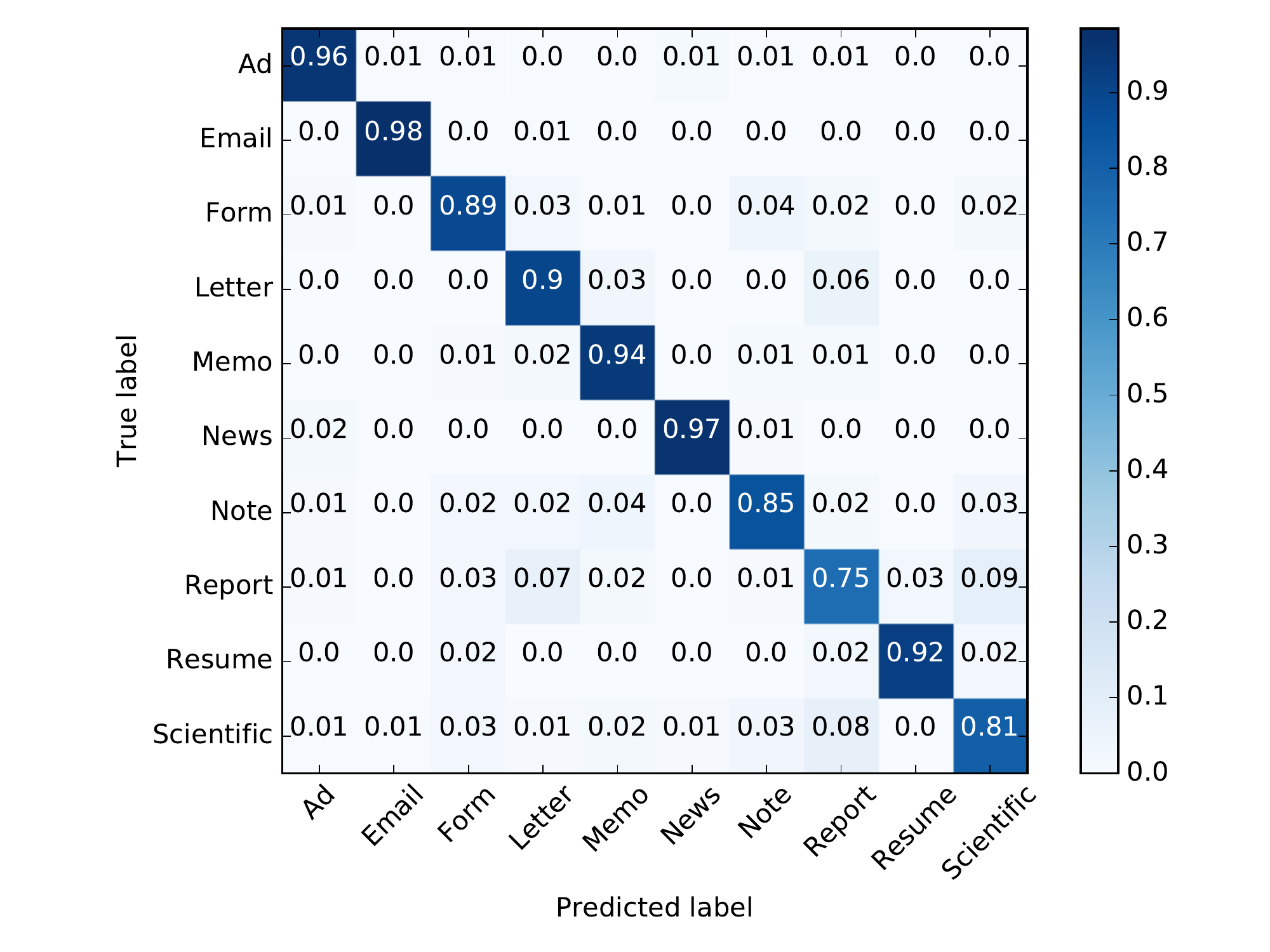}
        \caption{Confusion Matrix by a VGG-16 network trained on the Tobacco-3482 dataset.}
\label{fig:confusion}
\end{figure}

On average, VGG-16 performs very well on all experiments performed in this work. As can be seen in Fig.~\ref{fig:confusion} which shows the confusion matrix of a trained VGG-16 network, even the classes that were pointed out to be hard by Afzal et al.~\cite{afzal2015deepdocclassifier}, get significant performance boosts.

\begin{table}
\renewcommand{\arraystretch}{1.3}
\centering
\caption{Performance of the networks on the RVL-CDIP dataset with different weight initializations.}
\begin{tabular}{l|c|c}
 & ImageNet Pretraining & No Pretraining \\\hline
AlexNet & $88.60\,\%$ & $88.19\,\%$ \\\hline
GoogLeNet & $89.02\,\%$ & $88.60\,\%$ \\\hline
VGG-16 & $90.97\,\%$ & $89.41\,\%$ \\\hline
Resnet-50 & $90.40\,\%$ & $89.24\,\%$
\end{tabular}
\label{tab:accuracy_large}
\end{table}

\section{Conclusion and Future Work}


The outcome of the study brings insights both for the deep neural network architectures and the amount of required training data. The proposed approach of training on document images and then finetuning for a document based dataset improved the error by $60\,\%$.
We show that the random initialization performs worst and initialization based on document images performs best.
Furthermore, on the large-scale RVL-CDIP dataset, VGG-16 outperforms the other networks.
Finally, a relative error reduction of $11.5\,\%$ compared to the \sota is achieved. 
Future work may evaluate recurrent neural networks or a combination of convolutional and recurrent neural networks to improve the performance further.

\ifCLASSOPTIONcaptionsoff
  \newpage
\fi

\bibliographystyle{IEEEtran}
\bibliography{sample.bib}

\end{document}